\documentclass[letterpaper,10pt,conference]{ieeeconf}
\IEEEoverridecommandlockouts
\usepackage{amsmath}
\usepackage{amssymb}
\usepackage{bm}
\usepackage{booktabs}
\usepackage{graphicx}
\usepackage{wrapfig}
\usepackage{siunitx}
\usepackage{float}
\usepackage{hyperref}
\usepackage{float}
\usepackage{capt-of}
\usepackage{caption}
\usepackage{cite}

\usepackage{xcolor}

\definecolor{haozhipurpole}{HTML}{7A85C1}

\newcommand{\paragraphc}[1]{\vspace{0.2em}\noindent\textbf{#1}}

\newcommand{\website}{https://jessicayin.github.io/osmo_tactile_glove/}  % change
\definecolor{citecolor}{HTML}{0071bc}
\hypersetup{
    colorlinks=true,
    citecolor=citecolor,
    filecolor=citecolor,
    linkcolor=citecolor,
    urlcolor=citecolor
}

\title{\LARGE \bf OSMO: Open-Source Tactile Glove for Human-to-Robot Skill Transfer}

\author{Jessica Yin$^{1}$, Haozhi Qi$^{1,*}$, Youngsun Wi$^{1,2,*}$, Sayantan Kundu$^{1}$, Mike Lambeta$^{1}$, \\ William Yang$^{3}$, Changhao Wang$^{1}$, Tingfan Wu$^{1}$, Jitendra Malik$^{1}$, Tess Hellebrekers$^{1}$\vspace{0.06in}\\
$^1$Meta FAIR \, $^2$University of Michigan \, $^3$University of Pennsylvania
\thanks{${}^*$ Equal contribution.}
}

\begin{document}

\maketitle
\thispagestyle{empty}
\pagestyle{empty}

\begin{abstract}
Human video demonstrations provide abundant training data for learning robot policies, but video alone cannot capture the rich contact signals critical for mastering manipulation. We introduce OSMO, an open-source wearable tactile glove designed for human-to-robot skill transfer. The glove features 12 three-axis tactile sensors across the fingertips and palm and is designed to be compatible with state-of-the-art hand-tracking methods for in-the-wild data collection. We demonstrate that a robot policy trained exclusively on human demonstrations collected with OSMO, \textit{without any real robot data}, is capable of executing a challenging contact-rich manipulation task. By equipping both the human and the robot with the same glove, OSMO minimizes the visual and tactile embodiment gap, enabling the transfer of continuous shear and normal force feedback while avoiding the need for image inpainting or other vision-based force inference. On a real-world wiping task requiring sustained contact pressure, our tactile-aware policy achieves a 72\% success rate, outperforming vision-only baselines by eliminating contact-related failure modes. We release complete hardware designs, firmware, and assembly instructions to support community adoption.

\end{abstract}

\section{Introduction}
Tactile sensing enables humans to excel at manipulation by providing real-time feedback about contact forces that vision alone cannot capture. Consider trying to dice a carrot from video alone; one cannot observe the nuanced force control that makes the task successful. Many different applied forces can result in nearly identical visual appearances, leaving critical information about force control invisible to vision. This ambiguity poses a fundamental challenge for robots that depend on vision-only inputs. As a result, there is growing interest in capturing human tactile data to guide robots toward human-level dexterity~\cite{xudexumi,luo2024tactile,adeniji2025feel,martin2004tactile}.

Human datasets have emerged as a valuable resource for robotics research due to the ease and accessibility of data collection. The vast majority of human datasets consist of videos~\cite{zhan2024oakink2,fu2025gigahands, banerjee2025hot3d}. 
Recent progress in wearable devices for AR/VR and vision-based hand tracking~\cite{guzey2025dexterity, tao2025dexwild, bharadhwaj2024track2act} has further expanded the use of human videos as a data source for robots. However, video alone is often insufficient, as it critically lacks the tactile information essential for contact-rich manipulation tasks.

\begin{figure}[ht!]
\centering
\includegraphics[width=\columnwidth]{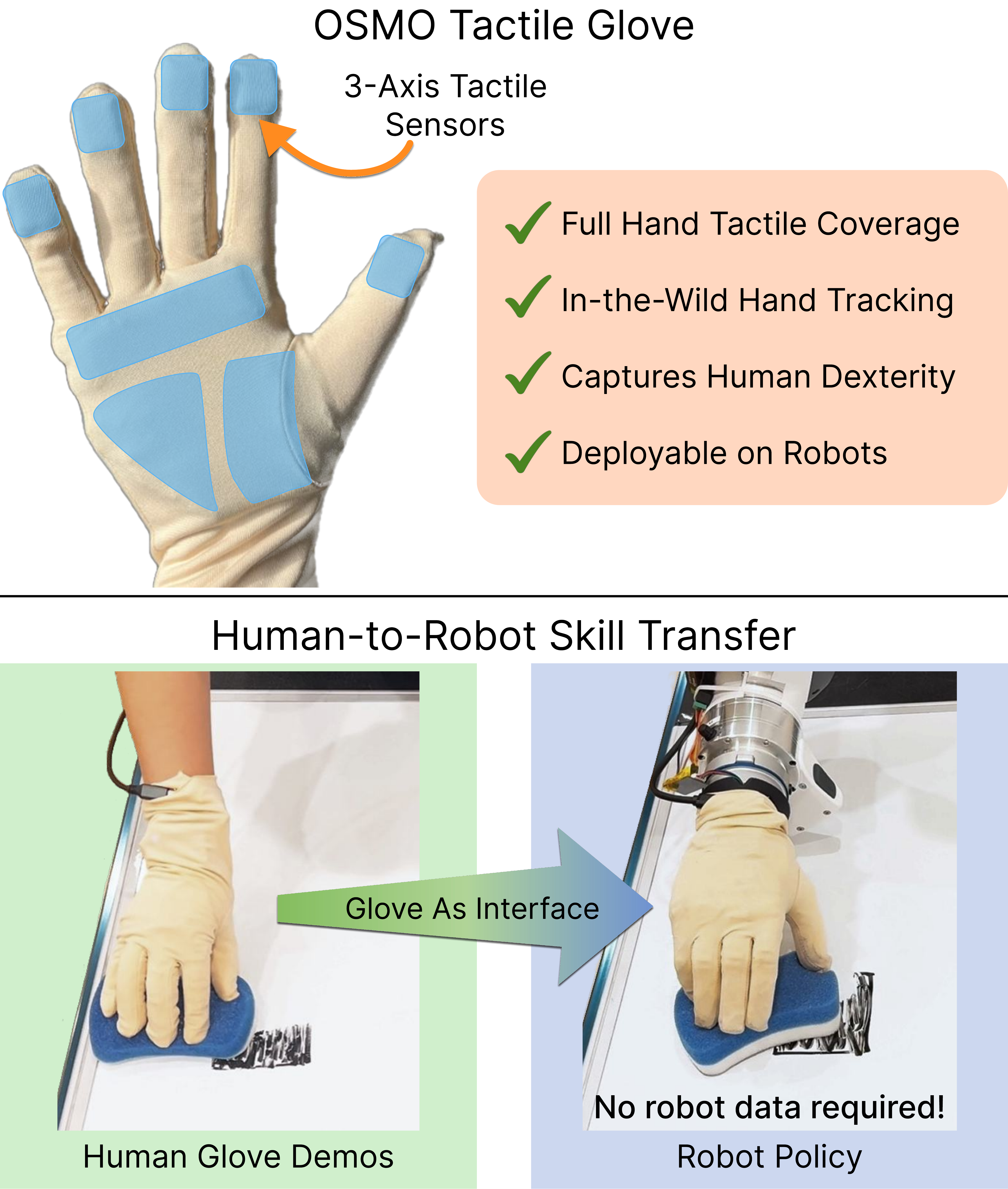}
\caption{(A) The OSMO tactile glove for collecting in-the-wild human demonstrations provides full-hand coverage with 3-axis tactile sensors, integrates seamlessly with hand-tracking systems, and can be directly deployed on robots. (B) We demonstrate that a contact-rich wiping policy trained solely on OSMO tactile-glove human demonstrations can be directly deployed on a robot, outperforming vision-only policies. More videos, code, and design files are available at \href{\website}{the project website}.}
\label{fig:teaser}
\vspace{-1.8em}
\end{figure}

To address this, we present \textbf{OSMO}: the \textbf{\underline{O}}pen \textbf{\underline{S}}ource tactile glove for hu\textbf{\underline{M}}an-to-rob\textbf{\underline{O}}t skill transfer (Figure~\ref{fig:teaser}). We propose a thin, wearable tactile glove that enables in-the-wild human demonstrations while preserving natural interaction and capturing rich contact information. The glove incorporates magnetic tactile sensors distributed across the fingertips and palm to measure both shear and normal forces, and it is designed to minimize encumbrance while maximizing sensing coverage. OSMO is also broadly compatible with state-of-the-art hand trackers for capturing key hand-pose data, including: 1) wearable devices for egocentric video such as Aria 2 smart glasses\footnote{https://www.projectaria.com/} and Quest 3\footnote{https://www.meta.com/quest/quest-3/}, 2) the Manus Quantum hand tracking glove\footnote{https://www.manus-meta.com/products/quantum-metagloves}, which provides robustness to occlusions, and 3) off-the-shelf vision models (e.g., HaMeR~\cite{pavlakos2024reconstructing}, Dyn-HaMR~\cite{yu2025dynhamr}) for processing generic RGB videos.

We demonstrate the viability of the OSMO tactile glove for human-to-robot skill transfer and propose a pipeline for training robot policies directly from human tactile-glove demonstrations. Using OSMO as the shared interface, we bridge the visual-tactile gap between the human demonstrator and the robot by training a policy for a contact-rich manipulation task \textbf{using only human demonstrations, without any robot data}. We focus on a wiping task, which requires consistent, sustained pressure that is difficult to infer from visual observations alone. Although we evaluate our approach on the Psyonic Ability Hand, the method could naturally extend to other anthropomorphic robot hands, such as the Inspire Hand or Sharpa Hand.

In conclusion, our key contributions are:
\begin{itemize}
    \item We introduce and open-source the OSMO tactile glove for collecting in-the-wild human demonstrations. We present the hardware design, sensor crosstalk mitigation strategies, and fabrication process. All design files are available at \href{\website}{the project website} to support broader community use and exploration of human tactile data.
    \item We develop an algorithmic pipeline that trains and deploys contact-rich robot policies \emph{using only human demonstrations} collected with the OSMO glove. The resulting tactile-aware policy significantly outperforms vision-only baselines on a wiping task that requires sustained contact.
\end{itemize}

\section{Related Work}

\paragraphc{Tactile Sensors for Robotics.}
Tactile sensing technologies offer distinct trade-offs in resolution and form factor. Optical tactile sensors \cite{lambeta2024digit360,lambeta2020digit,yin2022multimodal,yuan2017gelsight, sipos2024gelslim,donlon2018gelslim,taylor2022gelslim,romero2024eyesight,zhao2023gelsight} are the most common; they achieve high spatial resolution by embedding cameras to observe a deformable gel, deriving shear and normal forces from optical flow. While these sensors can be integrated into robot fingertips, they are bulky and difficult to adapt for varied surface geometries. Resistive tactile sensors \cite{zhu2025touch,huang20253d} offer thin, flexible form factors but typically measure only normal forces, missing the shear forces that are critical to understanding contact. Magnetic tactile sensors provide another skin-like alternative \cite{bhirangi2021reskin,bhirangi2025anyskin,tomo2017covering}. These sensors can capture shear and normal forces through magnetometer measurements of magnetic particle displacement. Their MEMS-based architecture naturally adapts to different form factors and surfaces. Our work builds on this line of magnetic sensing. Unlike prior work, which used single sensor-magnet pairs rigidly affixed to the robot, we integrate 12 distributed sensor-magnet pairs within a wearable glove and introduce crosstalk mitigation techniques to address noise introduced by this dense sensor array.

\begin{figure}[h!]
\centering
\includegraphics[width=\columnwidth]{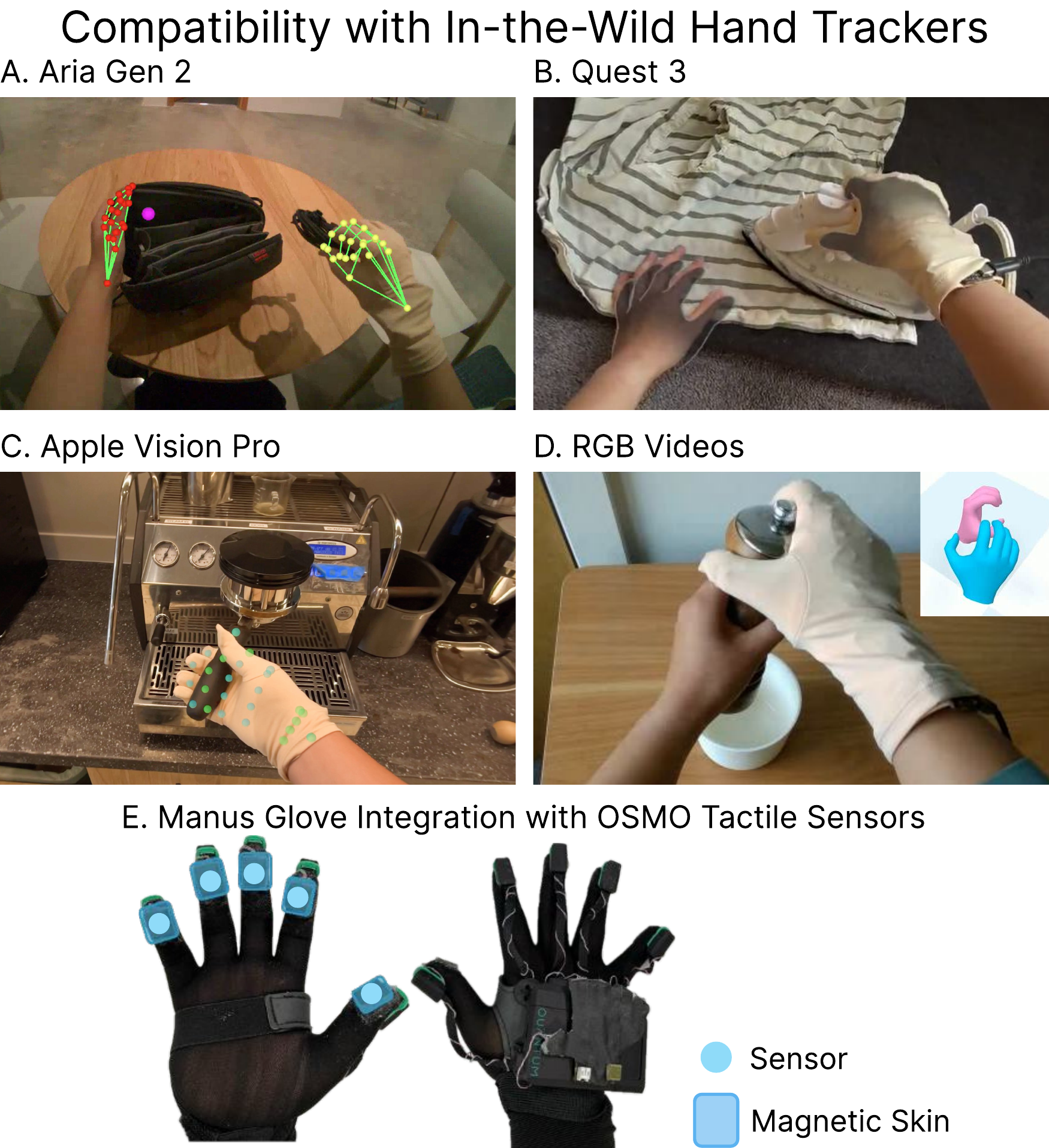}
\caption{The OSMO tactile glove is designed to be compatible with off-the-shelf hand-tracking devices for in-the-wild data collection. OSMO is worn on the right hand, and the device’s native hand-tracking estimates are overlaid. (A) Aria Gen 2, pick and place. (B) Quest 3, ironing. (C) Apple Vision Pro, espresso machine operation. (D) RGB video, pepper-grinder use. (E) Manus glove integration for hand-pose tracking robust to visual occlusion. The OSMO magnetic tactile sensors do not interfere with Manus finger tracking.}
\label{fig:hand_tracking}
\vspace{-1.6em}
\end{figure}

\paragraphc{Wearable Devices for Data Generation.}
Wearable devices have emerged as a powerful tool for robot learning by enabling more natural demonstration collection. Some works attach robot sensors directly to human fingertips \cite{yu2025mimictouch, adeniji2025feel, chen2025dexforce}, enriching demonstrations with tactile feedback but constraining manipulation to simple parallel-jaw grasps. For multi-fingered systems, exoskeleton gloves with integrated tactile sensors \cite{xudexumi, fang2025dexop, zhang2025doglove} address kinematic retargeting by mechanically constraining human motion to match robot degrees of freedom and linkage geometry. While this reduces the kinematic gap, each exoskeleton is specific to the robot design and thus requires a new one for each target robot hand. Flexible glove-based devices preserve full human dexterity, but they either provide kinematics-only \cite{tao2025dexwild,yin2025geometric,yin2025dexteritygen} or are limited to normal forces only \cite{SSundaram:2019:STAG, luo2024tactile}. Our tactile glove is the first flexible platform to capture both shear and normal forces while preserving full human dexterity.

\paragraphc{Learning from Human Demonstration.} 
Recent work leverages human video by learning transferable representations through latent skill spaces \cite{kim2025uniskill, wang2023mimicplay, yelatent}. These approaches effectively capture high-level skill abstractions, but fall short at low-level motor control and thus require additional robot data. Other studies learn explicit representations for hand-object interactions such as affordances \cite{liu2025egozero, sunil2025reactive, bahl2022human, goyal2022human}, dense point tracking \cite{bharadhwaj2024track2act, bharadhwaj2024gen2act, ren2025motion}, or hand-object masks \cite{bharadhwaj2024towards}. These methods produce transferable action representations that robots can potentially leverage, but mapping them to executable robot actions remains non-trivial. There are also methods that directly couple human video with robot policy learning by extracting kinematic action labels \cite{kareeregomimic,liokami, shi2025zeromimic, guzey2025dexterity, pan2025spider}, which demonstrate successful policy deployments for a variety of tasks. However, these vision-based approaches fundamentally cannot capture force information for contact-rich manipulation. Our work addresses this gap by collecting and transferring tactile feedback from human demonstrations, enabling force-aware policies unavailable from video alone.

\begin{figure}[t!]
\centering
\includegraphics[width=\columnwidth]{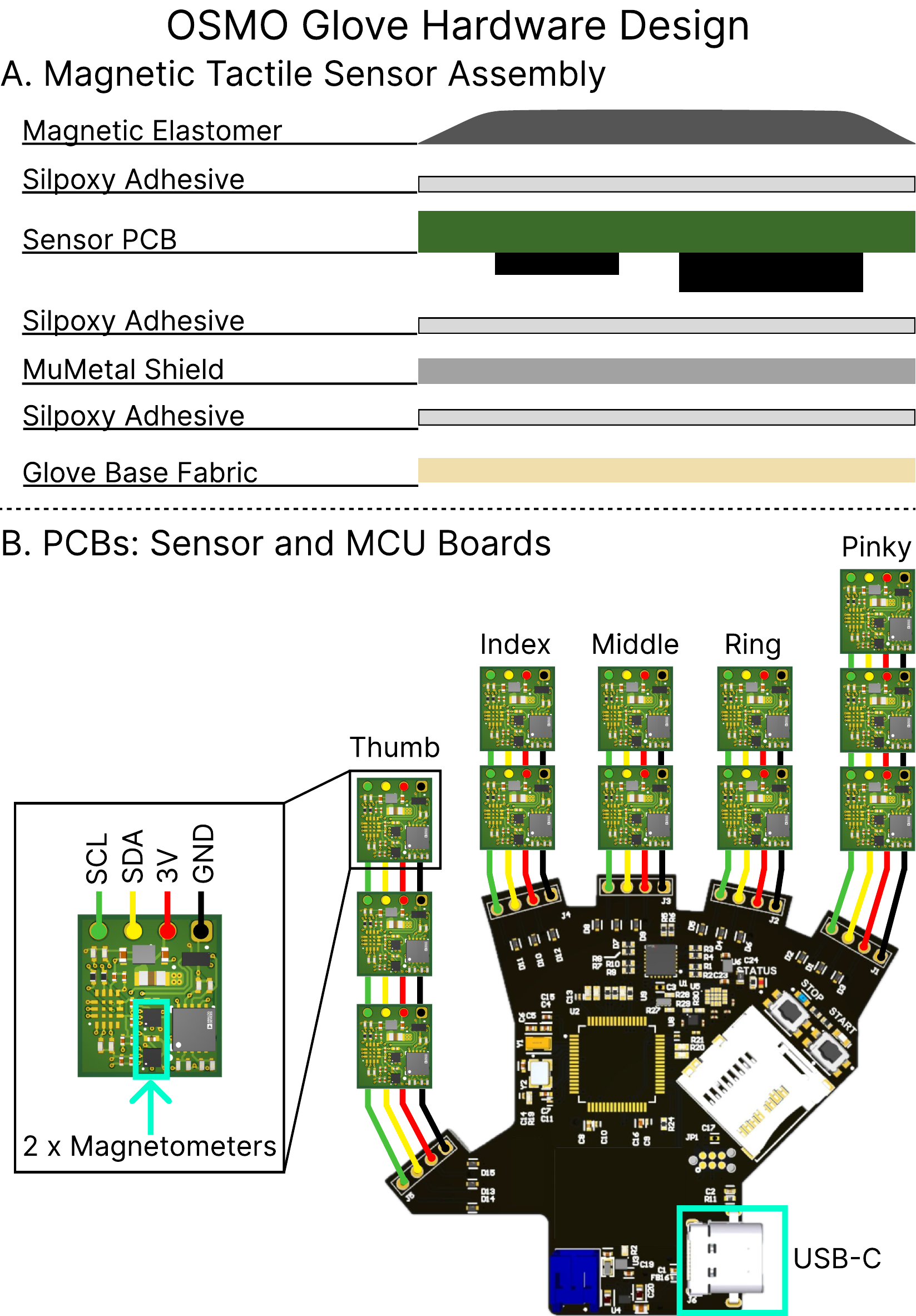}
\caption{(A) Layer-by-layer breakdown of the magnetic sensor assembly, including MuMetal shielding designed to reduce crosstalk and improve signal fidelity. (B) Wiring diagram of the sensors and microcontroller boards. Data is output via USB-C.}
\label{fig:hardware}
\vspace{-1.6em}
\end{figure}

\section{OSMO Tactile Glove Hardware Design}
We design OSMO with the following goals:

\paragraphc{1. Unrestrained human dexterity during demonstration collection.} Constraints on dexterity can create practical barriers to the speed and required skill level of collecting data. Recent advances in dexterous robot hand hardware~\cite{zorin2025ruka, shaw2025leap} and human-to-robot kinematic retargeting methods \cite{pan2025spider, adeniji2025feel, yin2025geometric} can enable direct transfer from human demonstrations to deployable robot policies, removing the need to restrain human dexterity during data collection.

\paragraphc{2. Rich normal and shear force sensing.} Human manipulation involves both normal and shear force feedback. Prior wearable tactile sensors have predominantly focused on normal forces \cite{murphy2025fits, luo2024adaptive}, limiting the contact information collected. Magnetic tactile sensors enable measurement of all three force axes in a thin form factor that allows for feel-through haptic feedback to the demonstrator. We equip the glove with 12 taxels on the fingertips and palm, with a sensing range of 0.3 N - 80 N.

\paragraphc{3. Full hand tactile coverage.} Contact during dexterous manipulation occurs across the entire hand surface, with the palm playing a critical role in stabilization and grasping \cite{bullock2011classifying}. We provide tactile coverage on both fingertips and palm while omitting the intermediate and proximal phalanges. This design choice accommodates varying hand sizes by allowing flexible finger lengths while ensuring consistent sensor placement at the primary contact regions. The palm sensors are arranged in three sections aligned with the major planar regions of palm motion.

\paragraphc{4. Broad compatibility with in-the-wild hand tracking methods.} Human demonstrations require hand and wrist poses to provide action labels. Recent advances in vision-based hand estimation with wearable headsets and RGB cameras enable tracking in the wild \cite{yu2025dynhamr, pavlakos2024reconstructing, guzey2025dexterity}, but these methods are trained primarily on bare human hands. We design the OSMO glove's visual appearance to minimize the distribution shift from bare hand datasets: a low-profile electronics layout and beige-colored fabric for compatibility with off-the-shelf vision-based trackers (Figure \ref{fig:hand_tracking}A-C). For demonstrations requiring occlusion-robust tracking, the magnetic sensors can be integrated with the Manus glove (Figure \ref{fig:hand_tracking}D) without electromagnetic interference, as the OSMO tactile sensors operate at a different frequency.

\paragraphc{5. Deployable on both human and robot hands.} Embodiment gaps between demonstration and deployment hardware can make policy transfer more difficult \cite{xudexumi, tao2025dexwild}. To minimize this gap, we design the OSMO glove with a stretchable base layer and sensor layout that can be worn by both human demonstrators and biomimetic robot hands, such as the Psyonic Ability Hand. A shared glove platform can help minimize the gap in tactile and visual feedback at data collection and deployment time.

These design objectives represent our vision for tactile glove hardware to study human-to-robot skill transfer. We present the OSMO glove hardware design to realize these goals while balancing practical trade-offs in accessibility and ease of fabrication.

\subsection{Magnetic Tactile Sensors}

\paragraphc{Sensing Principle.} The sensors build on previous magnetic tactile work \cite{bhirangi2021reskin, bhirangi2025anyskin}, while extending it for a wearable glove form-factor. The sensing principle relies on the changes in the relative position between magnetometers and nearby soft magnetic elastomers. When the soft magnetic elastomers are deformed, the magnetometers measure changes in magnetic flux, which can be correlated to applied XYZ forces. In this work, we directly use the raw magnetic flux signals ($\mu$T) as tactile data. 

\paragraphc{Assembly.} The soft magnet patches are cast in 3D-printed molds with a silicone (EcoFlex 00-30; Smooth-on). Magnetic microparticles (MQFP-15-7; Magnequench) are mixed with the two-part pre-polymer in a 1:1:1 ratio. After curing at room temperature, the patches are demolded and axially magnetized in a pulse magnetizer for 8 seconds at $2\text{ kV}$. Each individual sensor PCB consists of two 3-axis magnetometers (BMM350; Bosch) and one 6-axis IMU (BHI360; Bosch). The magnetic patches are directly adhered to the base of the PCB with Silpoxy (Smooth-on), where the flat surface ensures more consistent deformation of the soft magnet patch. The MuMetal magnetic shield (11 mm $\times$ 12 mm $\times$ 0.1524 mm) is cut via waterjet and lightly sanded prior to adhesion to the glove base fabric and the sensor PCB with Silpoxy. The sensors are wired in series with tinsel wire to the shared microcontroller (STM32; ST Microelectronics), which aggregates all the incoming data from the sensors to an SD card or host computer via USB-C. The glove PCB designs and layer-by-layer sensor assembly are shown in Figure \ref{fig:hardware}A-B.

\begin{figure}[t!]
\centering
\vspace{0.5em}
\includegraphics[width=\columnwidth]{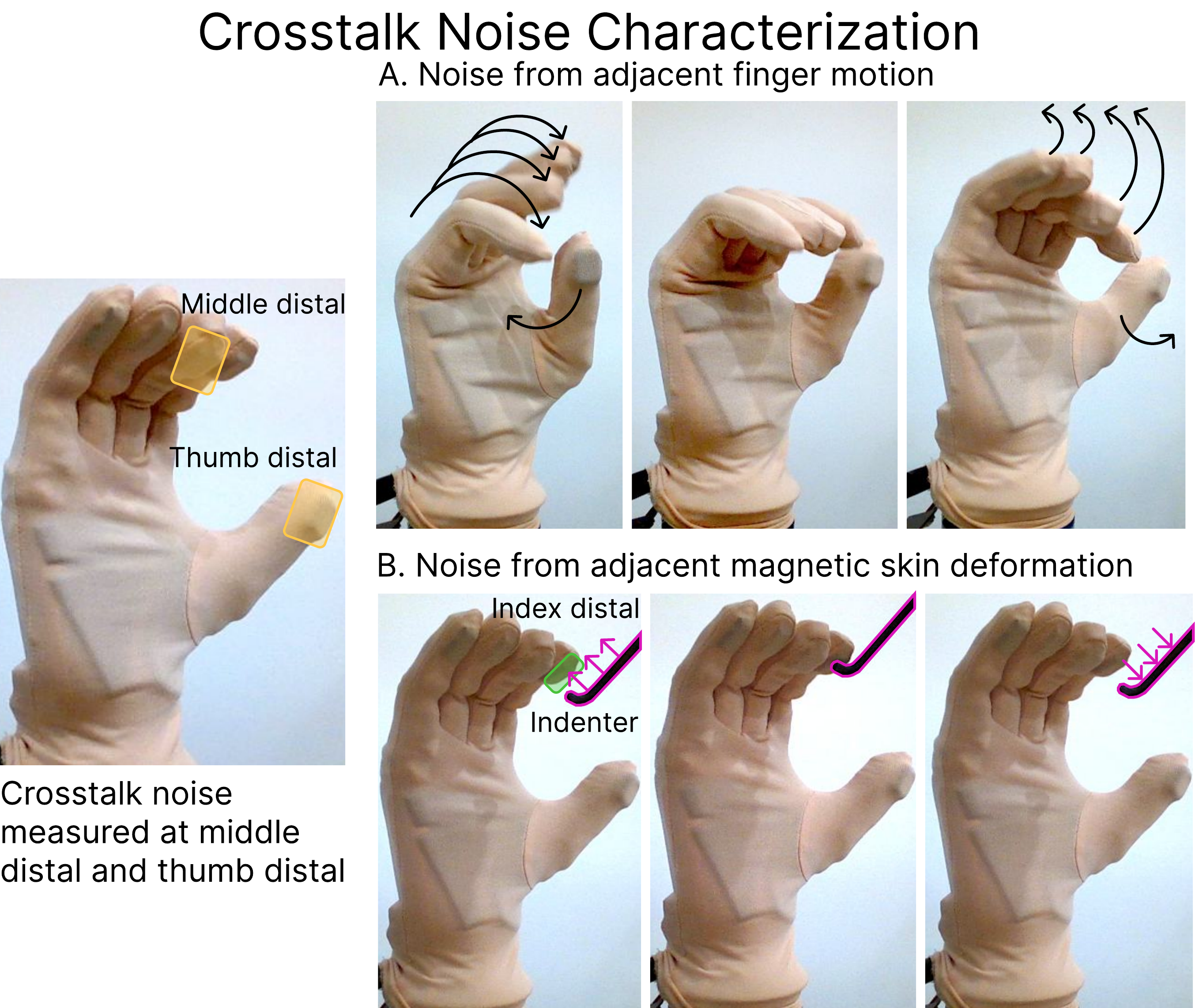}
\caption{Crosstalk characterization experiments with the glove worn by a robot hand. (A) Finger motion: the robot executes a sinusoidal finger wave. (B) Adjacent contact deformation: the soft magnetic patch on the index distal phalanx is repeatedly pressed.}
\label{fig:cross_talk}
\vspace{-0.5em}
\end{figure}

\begin{table}[t!]
\centering
\caption{RMS crosstalk noise ($\mu$T) across magnetometer configurations: unshielded single magnetometer (baseline), unshielded with 2 magnetometers (differential sensing), and shielded with 2 magnetometers (our approach). Lower values indicate better crosstalk suppression.}
\label{tab:cross_talk}
\begin{tabular}{lcccc}
\toprule
\multicolumn{5}{c}{\textbf{Adjacent Finger Motion - RMS Noise (µT)}} \\
\midrule
 \textbf{Thumb Distal} & \textbf{X-Axis} & \textbf{Y-Axis} & \textbf{Z-Axis} & \textbf{Avg  $\downarrow$} \\
\midrule
Unshielded + 1 mag & 205.3 & 262 & 2.9 & 156.73 \\
Unshielded + 2 mags & 193.7 & 237.2 & 4.01 & 144.97 \\
Shielded + 2 mags (Ours) & 31.39 & 46.63 & 143 & 73.67 \\
\midrule
 \textbf{Middle Distal} & \textbf{X-Axis} & \textbf{Y-Axis} & \textbf{Z-Axis} & \textbf{Avg  $\downarrow$} \\
\midrule
Unshielded + 1 mag & 374.9 & 69.35 & 1.371 & 148.54 \\
Unshielded + 2 mags & 67.48 & 38.64 & 1.268 & 35.80 \\
Shielded + 2 mags (Ours) & 18.1 & 20.21 & 57.33 & 31.88 \\
\toprule
\multicolumn{5}{c}{\textbf{Adjacent Contact at Index Distal - RMS Noise (µT)}} \\
\midrule
 \textbf{Thumb Distal} & \textbf{X-Axis} & \textbf{Y-Axis} & \textbf{Z-Axis} & \textbf{Avg  $\downarrow$} \\
\midrule
Unshielded + 1 mag & 2.117 & 4.356 & 0.7628 & 2.412 \\
Unshielded + 2 mags & 1.824 & 12.49 & 0.6617 & 4.992 \\
Shielded + 2 mags (Ours) & 0.5289 & 0.5242 & 3.113 & 1.389 \\
\midrule
\textbf{Middle Distal} & \textbf{X-Axis} & \textbf{Y-Axis} & \textbf{Z-Axis} & \textbf{Avg  $\downarrow$} \\
\midrule
Unshielded + 1 mag & 78.9 & 34.95 & 0.7059 & 38.19 \\
Unshielded + 2 mags & 22.01 & 11.21 & 0.7154 & 11.31 \\
Shielded + 2 mags (Ours) & 7.324 & 16.03 & 31.24 & 18.20 \\
\bottomrule
\vspace{-3em}
\end{tabular}
\end{table}

\subsection{Multi-Sensor Crosstalk Mitigation}
Scaling from single sensor implementations to 12 closely-spaced magnetic sensors introduces significant crosstalk. Due to the close proximity, each magnetometer can detect magnetic field perturbations from relative motion and deformation of the adjacent soft magnets during finger and palm motion, which adds noise to the tactile signal. Prior magnetic tactile sensors \cite{bhirangi2021reskin, bhirangi2025anyskin, pattabiraman2025eflesh} have not addressed multi-sensor crosstalk at this scale in a wearable form factor. Here, we introduce two new techniques to our sensor design for crosstalk mitigation: 1) MuMetal shielding integrated into the sensor architecture to attenuate external magnetic fields, and 2) a dual magnetometer differential sensing layout per taxel to reduce common-mode noise. We evaluate the effectiveness of our mitigation strategies on scenarios that commonly generate crosstalk during glove operation: finger motion and adjacent contact deformation.

\paragraphc{Experimental Setup.} We measure crosstalk noise at two sensor locations expected to experience significant interference: the middle finger distal sensor (surrounded by four adjacent fingertip sensors) and the thumb distal sensor (the largest range of motion). Both experiments use the glove worn by a Psyonic Ability robot hand for consistency. We compare three sensor configurations: 1) unshielded with a single magnetometer (baseline), 2) unshielded with dual magnetometers (differential sensing only), and 3) shielded with dual magnetometers (our approach). Noise is quantified as root-mean-squared (RMS) magnetic field variation in microtesla ($\mu$T), where lower is better.

\paragraphc{Finger Motion.} The robot executes a programmed finger wave where all fingers move sequentially in a smooth sinusoidal pattern (Figure \ref{fig:cross_talk}A). During this motion, the sensors do not experience any contact, but they are exposed to magnetic field variations from the adjacent moving magnets and Earth's magnetic field. We record 5 trials of 60 seconds each from random initial positions and compute RMS noise across all three axes.

\paragraphc{Adjacent Contact Deformation.} The index distal soft magnetic patch is pressed 10 times while measuring crosstalk at the nearby middle and thumb distal sensors (Figure \ref{fig:cross_talk}B). This simulates realistic contact scenarios where deformation of one sensor's magnetic elastomer may affect neighboring magnetometers.

\paragraphc{Results.} Table \ref{tab:cross_talk} shows that our combined approach (shielded + 2 magnetometers) reduces average noise by 57\% compared to a single magnetometer and by 18\% compared to the dual-magnetometer, unshielded configuration across both experiments. The MuMetal shielding provides additional attenuation particularly in the X-Y plane, while we observe an increase in Z-axis noise with the shield, which is expected: the MuMetal layer attenuates in-plane magnetic flux but concentrates some field lines along the Z-axis. However, the average noise reduction across all axes justifies this trade-off. We note that the finger motion experiment includes approximately 20-60 $\mu$T of noise from motion through Earth's magnetic field \cite{carrigan1979magfield}, which is included in the total RMS noise metric. For additional context, a 1 N applied force will generate magnetic signals in the range of 300 $\mu$T.

\begin{figure}[t!]
\small
\centering
\includegraphics[width=\columnwidth]{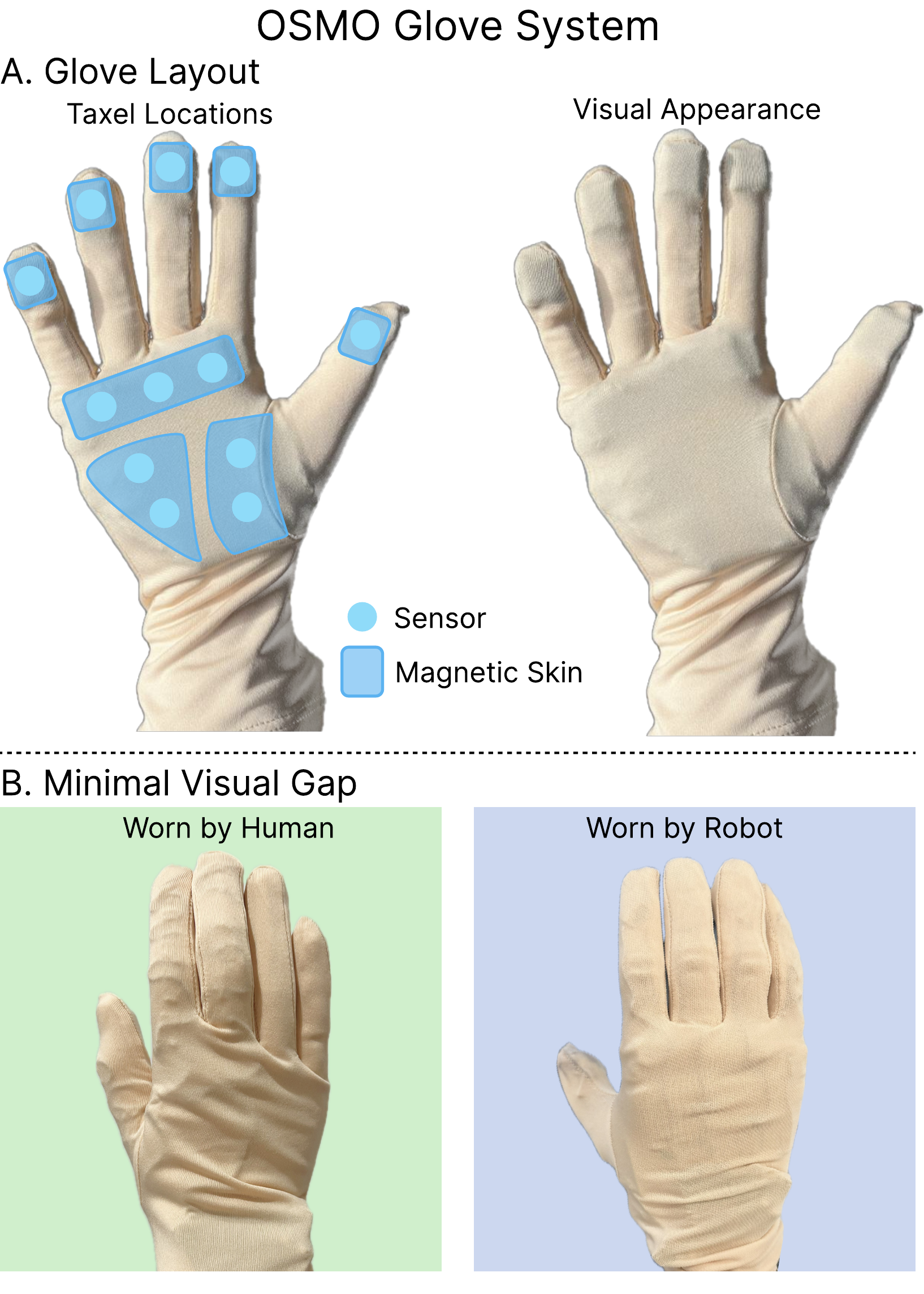}
\vspace{-1em}
\caption{A) Glove layout maximizing sensing coverage while minimizing encumbrance. B) The shared glove platform minimizes the visual gap between the human and robot hand. }
\label{fig:system}
\vspace{-2em}
\end{figure}

\subsection{System Integration}

\paragraphc{Electronics Interface.} The 12 sensor PCBs communicate with an STM32 microcontroller via I2C, enabling modular addition or removal of sensors. Custom firmware is flashed via STM32CubeIDE. Data streams to a host computer via USB-C, with native Python and ROS2 interfaces supporting time-synchronized sensor packets.

\paragraphc{Assembly.} Sensor PCBs are positioned at target locations with the microcontroller mounted at the wrist. Tinsel wires are routed in a serpentine pattern along the dorsal hand surface using Silpoxy adhesive, allowing flexible motion without straining solder connections. An outer beige glove provides uniform appearance and protects the electronics. The complete glove requires only USB-C for power and data, and accommodates both human and robot hands. For Manus glove integration, we follow the same procedure but route wiring alongside existing Manus cables and omit the outer beige glove layer.

\section{Glove2Robot Policy Pipeline}

We demonstrate human-to-robot skill transfer on a contact-rich wiping task using the OSMO tactile glove. Wiping requires maintaining consistent, sustained contact pressure, which tactile sensing can capture directly but vision-only can struggle to infer. For this task, we use the five fingertip tactile sensors and omit the palm sensors, as the palm does not provide task-relevant information. The same glove is used to collect human demonstrations and for policy deployment.

The OSMO tactile glove provides two key advantages that typically require substantial preprocessing. First, using the same glove on both the human and the robot greatly reduces the visual domain gap, enabling the use of an off-the-shelf vision encoder without the image editing, inpainting, or hand-masking steps commonly required to reconcile differences between human and robot embodiments. Second, the glove enables rich, continuous force transfer directly from human to robot: the robot policy utilizes continuous shear and normal tactile signals at deployment time, despite being trained entirely on human tactile data.

\subsection{Collecting Human Demonstrations}
We record demonstrations with the user wearing the tactile glove and one scene camera (Realsense D435) that captures IR and RGB images. All data streams are logged as ROS2 bag files at 25 Hz. We assume: 1) the camera intrinsics and extrinsics for both RGB and IR sensors are known, 2) the camera is extrinsically calibrated to the robot base frame, and 3) the human demonstrations take place within the robot's reachable workspace. No objects in the scene are explicitly tracked. In total, we record $140$ demonstrations, corresponding to approximately two hours of data.

\subsection{Processing Human Demonstration Data}
\paragraphc{Hand Estimation.} We estimate the human hand pose from the RGB images. A hand mask is first extracted using SAM2~\cite{ravi2024sam}, and the resulting mask is converted into a bounding box. HaMeR is then applied to the cropped hand image to obtain 3D keypoints, a wrist pose, and a full hand mesh. HaMeR’s predictions alone can be noisy due to depth ambiguity. To refine them, we compute stereo depth from the RealSense IR pair using FoundationStereo~\cite{wen2025foundationstereo} and combine the depth map with the SAM2 mask to form a hand point cloud. This point cloud is projected into the robot frame using calibrated extrinsics. Assuming the RGB image and point cloud are aligned, we identify the nearest neighborhood of points corresponding to the HaMeR wrist keypoint, and shift the HaMeR mesh to the median height of these points. Finally, fingertip and wrist trajectories are smoothed using a Savitzky-Golay filter~\cite{savitzky1964smoothing}. In the end, we have a clean and smoothed fingertip and wrist trajectory for retargeting.

\begin{figure}[h!]
\centering
\includegraphics[width=\columnwidth]{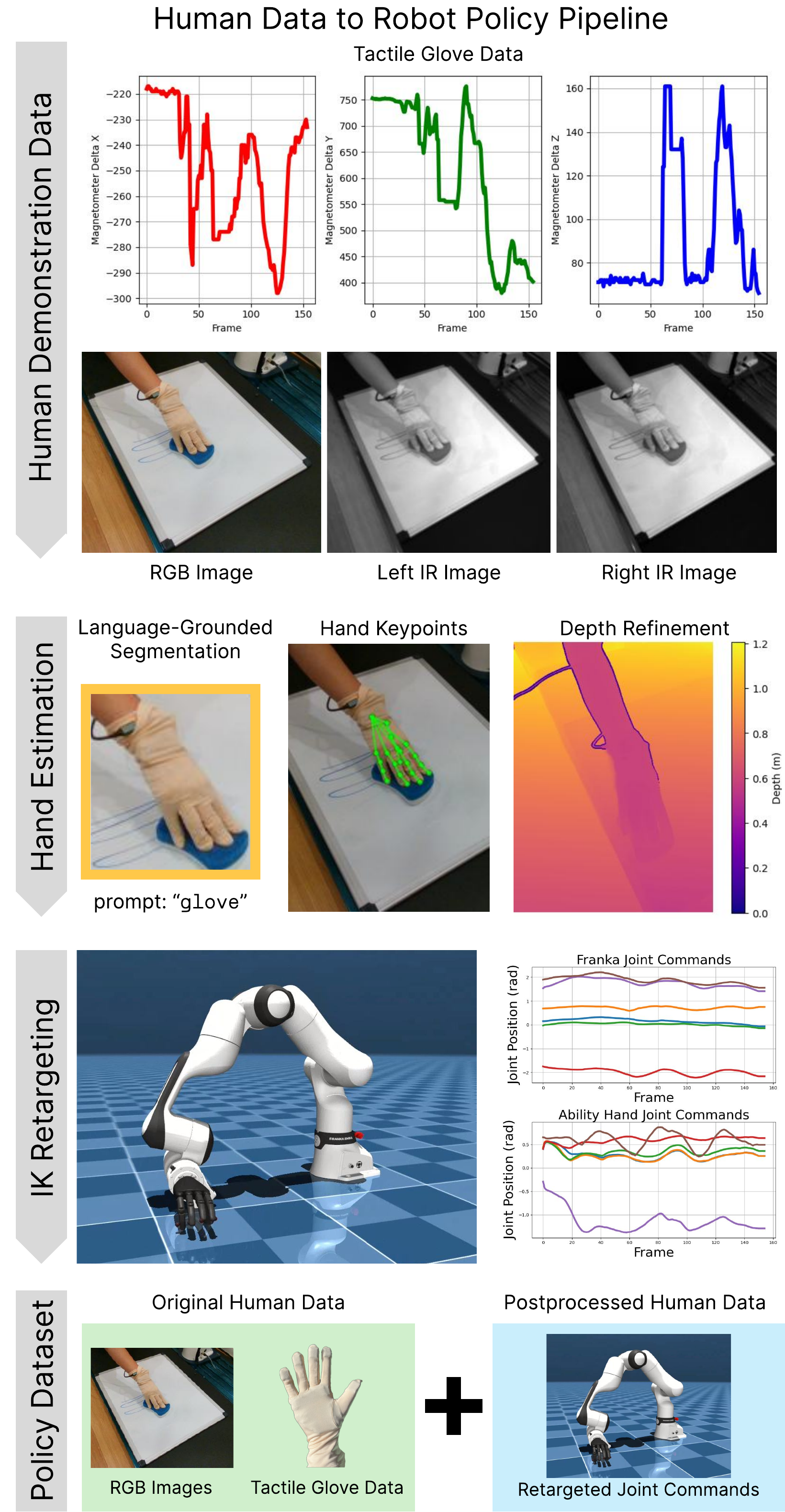}
\caption{Human demonstrations are captured with RGB + stereo IR cameras and OSMO tactile glove data. We estimate the hand pose and refine the global wrist position with depth. Inverse kinematics (IK) retargets the human hand positions to Franka arm and Ability Hand joint targets. The final dataset directly uses RGB images and tactile glove data from the human demonstration with the retargeted robot joint positions for policy training.}
\vspace{-2.3em}
\label{fig:postprocess}
\end{figure}

\paragraphc{Kinematics Retargeting.} We convert the human fingertip and wrist trajectories into robot joint commands using an off-the-shelf IK solver~\cite{Zakka_Mink_Python_inverse_2025} in MuJoCo. Because the Ability Hand is a similar size to a human hand, the human fingertip positions can be used directly as IK targets. The human wrist trajectory serves as the target for the Franka arm. Two safety constraints are applied in MuJoCo: collision detection and a wrist velocity limit. Since wiping is a contact-rich task and the hand pose estimation can be noisy, we disable collision checks for the Ability Hand fingertips and Franka wrist against the ground plane. If either of these conditions are violated, the previous robot pose is repeated and the current unsafe pose is skipped. This stage produces robot commands that are time-synchronized with the human demonstration videos and tactile data.

\paragraphc{Dataset Overview. }We define the original human dataset as $D^H = \{T_0, T_1, \ldots, T_{N-1}\}$ as a collection of $N$ trajectories, where each $T_i = \{F_0, F_1, \ldots, F_{M-1}\}$ has M frames. Each frame has the following aligned data:

\begin{equation}
    F_k = \{I^H_{\text{rgb}}, I^H_{\text{IR}_\text{left}}, I^H_{\text{IR}_\text{right}}, g^H\}
\end{equation}

where $I^H_{\text{rgb}} \in \mathbb{R}^{H \times W \times 3}$ is an RGB image and $I^H_{\text{IR}_\text{left}}, I^H_{\text{IR}_\text{right}} \in \mathbb{R}^{H \times W}$ are infrared images from the left and right sensors, respectively. The term $g^H \in \mathbb{R}^{3 \times 2 \times 5}$ represents tactile data from the glove, structured across the XYZ spatial dimensions, 2 magnetometers per sensor PCB, and 5 sensors for each of the glove fingertips. We directly use the magnetic field strength ($\mu$T) as tactile signals.

The final robot dataset is $D^R = \{T_0, T_1, \ldots, T_{N-1}\}$ with $N$ trajectories, where each $T_i = \{F_0, F_1, \ldots, F_{M-1}\}$ has M frames. Each frame has been transformed into

\begin{equation}
    F_k = \{I^H_{\text{rgb}}, q^R, g^H\}
\end{equation} 
where $I^H_{\text{rgb}} \in \mathbb{R}^{H \times W \times 3}$ is the source RGB image, $q^R \in \mathbb{R}^{13}$ is the robot's joint positions (7-DoF Franka + 6-DoF Ability Hand), and $g \in \mathbb{R}^{3 \times 2 \times 5}$ denotes the glove tactile data.

\subsection{Policy Training}

We show that the data collected with the OSMO tactile glove enables training an autonomous policy that successfully accomplishes the wiping task. Specifically, we learn a policy $\pi(a \mid F_k)$ that takes the RGB image, robot state, and tactile signals as inputs, and outputs an action chunk~\cite{zhao2023learning} consisting of target joint positions for both the arm and hand. 

\paragraphc{Policy Representation.} We represent this policy as a diffusion policy~\cite{chi2024diffusionpolicy}, which generates action sequences by denoising latent trajectories conditioned on observations. This formulation naturally captures the trajectory-level variability and multi-modality present in human demonstrations.

\paragraphc{Network Architecture.} The policy consists of two components. First, we encode each modality separately: we use two MLPs to process the robot states and tactile signals, and a DINOv2~\cite{oquab2023dinov2} encoder to extract image features. The resulting embeddings are concatenated to form the global conditioning vector. Second, a denoising network conditioned on this vector iteratively predicts the action sequence. We follow the standard diffusion policy architecture, using FiLM~\cite{perez2018film}-based conditioning and a U-Net~\cite{ronneberger2015u} denoising network.

Following~\cite{chi2024universal}, we use an observation horizon of 1 and predict the next 16 timesteps during training. During deployment, the policy executes the first 4 actions from each action chunk. We adopt the DDPM scheduler~\cite{ho2020denoising} with 100 denoising steps during both training and deployment.

\paragraphc{Data Processing.} Following~\cite{barreiros2025careful}, we normalize the robot state and tactile signals by computing the 2nd and 98th percentiles for each channel, linearly scaling values into this range, and clipping them to $[-1.5, 1.5]$:
\begin{equation}
y_i = \text{min}(\text{max}(-1.5, 2 \frac{x_i - x^{0.02}}{x^{0.98} - x^{0.02}}), 1.5).
\end{equation}

In addition, for tactile sensing, we apply differential sensing by subtracting paired magnetometer readings on each sensor PCB at every timestep before passing them through the MLP encoder.

For RGB images, we crop them to a fixed resolution before feeding them to the DINOv2 encoder. During training, they are randomly cropped to $224\times 224$, and during testing, they are center cropped to the same resolution.

\paragraphc{Training.} We optimize the policy using Adam~\cite{kingma2014adam}. Unless otherwise specified, we keep the DINOv2 image encoder frozen during training. We use a learning rate of 0.0002, a batch size of 128, and train for 2000 epochs.

\begin{figure}[!t]
\centering
\includegraphics[width=\columnwidth]{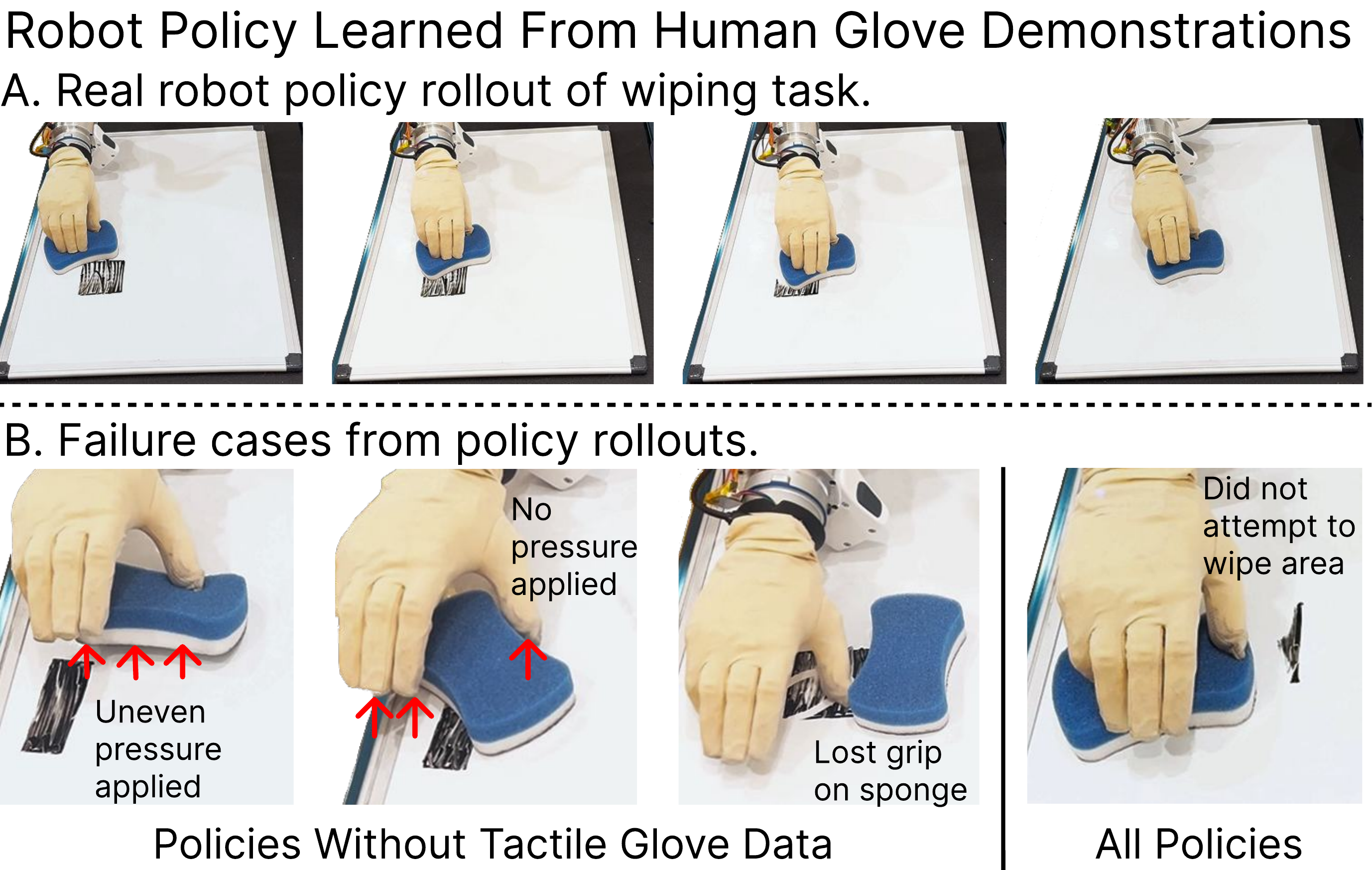}
\caption{(A) Real robot policy rollouts with the Psyonic Ability Hand and Franka robot arm. (B) Failure modes for policies with and without tactile feedback.}
\label{fig:policy}
\vspace{-1em}
\end{figure}

\begin{table}[t]
\normalsize
\centering
\caption{Wiping policy performance for 12 rollouts, where the success metric is percentage of marker pixels erased.}
\label{tab:policy_perf}
\begin{tabular*}{\columnwidth}{@{\extracolsep{\fill}}lcc@{}}
\toprule
\textbf{Sensing Modalities} & \textbf{Success (\%)} \\
\midrule
Proprioception & 27.12 $\pm$ 32.38 \\
Vision + proprioception & 55.75 $\pm$ 30.01 \\
\textbf{Tactile glove + vision + proprio.} & \textbf{71.69 $\pm$ 27.43} \\
\bottomrule
\end{tabular*}
\vspace{-1.5em}
\end{table}

\subsection{Policy Evaluation}
\paragraphc{Experimental Setup.} We deploy policies in the same environment used to collect the human demonstrations, maintaining identical camera placement and using the same OSMO glove on the robot (Figure \ref{fig:policy}A). The robot initializes to a fixed pose with the sponge in contact with the fingertips. Vision and tactile inputs stream at 25 Hz while the diffusion policy outputs action sequences at 2 Hz. We evaluate each policy variant across 12 rollouts with two distinct marker patterns positioned at different locations on the whiteboard. Marker patterns are drawn with stencils for consistency in shape and position. Each rollout runs for 90 s. 

\paragraphc{Baselines.} We compare our proposed policy (tactile glove, proprio., and vision) against two ablations: proprio-only, proprio. and vision. 

\paragraphc{Success Metric.} Success is measured as the percentage of marker pixels erased, computed by comparing binarized images of the initial and final marker states.

\paragraphc{Results and Analysis.} Table \ref{tab:policy_perf} shows that tactile feedback significantly improves task performance. Our tactile policy achieves 71.69\% average success, outperforming vision+proprio and proprio-only baselines. We qualitatively analyze the rollouts and find distinct causes of lower success rates across the different policies. All policies suffer from not attempting to wipe the entire marker pattern, likely due to accumulated position error from human hand pose estimation, kinematic retargeting, and camera calibration. However, policies without tactile feedback frequently exhibit contact-related failures: inconsistent or insufficient applied pressure on the sponge or losing grasp of the sponge. The elimination of contact-related failures and improved success rate show that the tactile glove feedback is beneficial for this contact-rich task.

\section{Conclusion}
We present the OSMO tactile glove, an open-source glove system to study human-to-robot skill transfer. The glove integrates 12 magnetic tactile sensors across the fingertips and palm for shear and normal force sensing, leveraging MuMetal shielding and differential sensing to mitigate crosstalk in our dense multi-sensor layout. The OSMO glove achieves broad compatibility with in-the-wild hand tracking methods including wearable AR/VR headsets, RGB videos, and the Manus glove. We demonstrate that policies trained solely on human demonstrations with the OSMO glove successfully transfer continuous tactile feedback and outperform vision-only baselines by eliminating contact-related failures. The shared glove platform between human demonstrator and robot deployment minimizes the visual domain shift, avoiding the need for image inpainting. By open-sourcing all hardware and software in this work, we aim to lower the barrier to collect tactile demonstrations and support community adoption.
\subsection{Limitations and Future Work} 
While the OSMO glove demonstrates effective human-to-robot skill transfer, several directions are promising for future exploration.

\paragraphc{Hardware Improvements.} First, the current sensor PCBs include onboard IMUs that remain unexplored in this work. We observe that vision-based hand tracking accuracy degrades significantly when fingertips are occluded, which occurs frequently during common tasks. OSMO's fingertip and palm-mounted IMUs could provide key information about relative motion through occlusion for fusion with vision to refine hand pose estimates. Second, fingertip spatial resolution could be improved through denser sensor layouts and wraparound coverage. The current design features one taxel on each fingertip's planar surface, but multi-taxel fingertip resolution could enable fine-grained contact position sensing, which can be particularly useful for precise tasks.

\paragraphc{Algorithmic Extensions.} Our task and evaluation focused on a unimanual task with fairly limited dexterity. Future work should explore tasks that require multi-finger coordination and bimanual tasks with two tactile gloves. Additionally, we can explore how to leverage tactile feedback to improve policy generalization through visual variations in the object and environment, potentially resulting in improvements in data efficiency.

\section*{Acknowledgment}
We thank Alexa Greenberg, Irmak Guzey, and Haritheja Etukuru for helping with hand-tracking demos. We thank Victoria Rose Most and her team for help with magnetic elastomer fabrication and Mason Greer for helpful discussions about the magnetometer sensors.

\bibliographystyle{IEEEtran}
\bibliography{references_arxiv}

\end{document}